# BioRED: A Rich Biomedical Relation Extraction Dataset

Ling Luo[1,+] , Po-Ting Lai[1,+] , Chih-Hsuan Wei[1,+] , Cecilia N Arighi[2] and Zhiyong Lu[1,*]

[1]National Center for Biotechnology Information (NCBI), National Library of Medicine (NLM), National Institutes of Health (NIH), Bethesda, MD 20894, USA, [2]University of Delaware, Newark, DE 19716, USA

*To whom correspondence should be addressed.

+ The authors wish it to be known that, in their opinion, the first three authors should be regarded as joint First Authors.

## Abstract

Automated relation extraction (RE) from biomedical literature is critical for many downstream text mining applications in both research and real-world settings. However, most existing benchmarking datasets for biomedical RE only focus on relations of a single type (e.g., protein-protein interactions) at the sentence level, greatly limiting the development of RE systems in biomedicine. In this work, we first review commonly used named entity recognition (NER) and RE datasets. Then we present BioRED, a first-of-its-kind biomedical RE corpus with multiple entity types (e.g., gene/protein, disease, chemical) and relation pairs (e.g., gene-disease; chemical-chemical) at the document level, on a set of 600 PubMed abstracts. Further, we label each relation as describing either a novel finding or previously known background knowledge, enabling automated algorithms to differentiate between novel and background information. We assess the utility of BioRED by benchmarking several existing state-of-the-art methods, including BERT-based models, on the NER and RE tasks. Our results show that while existing approaches can reach high performance on the NER task (F-score of 89.3%), there is much room for improvement for the RE task, especially when extracting novel relations (F-score of 47.7%). Our experiments also demonstrate that such a rich dataset can successfully facilitate the development of more accurate, efficient, and robust RE systems for biomedicine.

## 1   Introduction

Biomedical natural language processing (BioNLP) and text-mining methods/tools make it possible to automatically unlock key information published in the medical literature, including genetic diseases and their relevant variants [1, 2], chemical-induced diseases [3], and drug response in cancer [4]. Two crucial and building block steps in the general biomedical information extraction pipeline, however, remain challenging. The first is named entity recognition and linking (NER/NEL), which automatically recognizes the boundary of the entity spans (e.g., ESR1) of a specific biomedical concept (e.g., gene) from the free text and further links the spans to the specific entities with database identifiers (e.g., NCBI Gene ID: 2099). The second is relation extraction (RE), which identifies an entity pair with certain relations.

To facilitate the development and evaluation of NLP and machine learning methods for biomedical NER/NEL and RE, significant efforts have been made on relevant corpora development [5-10]. However, most existing corpora focus only on relations between two entities and within single sentences. For example, Herrero-Zazo et al. [8] developed a drug-drug interaction (DDI) corpus by annotating relations only if both drug names appear in the same single sentence. As a result, multiple individual NER/RE tools need to be created to extract biomedical relations beyond a single type (e.g., extracting both DDI and gene-disease relations).

Additionally, in the biomedical domain, extracting novel findings that represent the fundamental reason why an asserted relation is published as opposed to background or ancillary assertions from the scientific literature is of significant importance. To the best of our knowledge, none of the previous works on (biomedical) relation annotation, however, included such a novelty attribute.

In this work, we first give an overview of NER/NEL/RE datasets, and show their strengths and weaknesses. Furthermore, we present BioRED, a rich biomedical relation extraction dataset. We further annotated the relations as either novel findings or previously known background knowledge. We summarize the unique features of the BioRED corpus as follows: (1) BioRED consists of biomedical relations among six commonly described entities (i.e., gene, disease, chemical, variant, species, and cell line) in eight different types (e.g., positive correlation). Such a setting supports developing a single general-purpose RE system in biomedicine with reduced resources and improved efficiency. More importantly, several previous studies have shown that training a machine-learning algorithm on multiple concepts simultaneously on one dataset, rather than multiple single-entity datasets, can lead to better performance [11-13]. We expect similar outcomes with our dataset for both NER and RE tasks. (2) The annotated relations can be asserted either within or across sentence boundaries. For example, as shown in Figure 1 (relation R5 in pink), the variant "D374Y" of the PCSK9 gene and the causal relation with the disease "autosomal dominant hypercholesterolemia" are in different sentences. This task therefore requires relations to be inferred by machine reading across the entire document. (3) Finally, our corpus is enriched with novelty annotations. This novel task poses new challenges for (biomedical) RE research and enables the development of NLP systems to distinguish between known facts and novel findings, a greatly needed feature for extracting new knowledge and avoiding duplicate information towards the automatic knowledge construction in biomedicine.

**Figure 1.** An example of a relation and the relevant entities displayed on TeamTat (https://www.teamtat.org).

To assess the challenges of BioRED, we performed benchmarking experiments with several state-of-the-art methods, including BERT-based models. We find that existing deep-learning systems perform well on the NER task but only modestly on the novel RE task, leaving it an open problem for future NLP research. Furthermore, the detailed analysis of the results confirms the benefit of using such a rich dataset towards creating more accurate, efficient, and robust RE systems in biomedicine.

## 2   Overviews of NER/NEL/RE datasets

### 2.1 Named entity recognition and linking

Existing NER/NEL datasets cover most of the key biomedical entities, including gene/proteins [14-16], chemicals [17, 18], diseases [9, 19], variants [20-22], species [23, 24], and cell lines [25]. Nonetheless, NER/NEL datasets usually focus on only one concept type; the very few datasets that annotate multiple concept types [26, 27] do not contain relation annotations. Table 1 summarizes some widely used gold standard NER/NEL datasets including the annotation entity type, corpus size and the task applications.

**Table 1.** Overview of gold standard NER/NEL datasets.

| Dataset | Text size | Entity type (#mentions) | Task type |
| --- | --- | --- | --- |
| JBLPBA [26] | 2,404 abstracts | Protein (35,336), DNA (10,589), RNA (1,069), cell line (4,330) and cell type (8,639) | NER |
| NCBI Disease [19] | 793 abstracts | Disease (6,892) | NER, NEL |
| CHEMDNER [18] | 10,000 abstracts | Chemical (84,355) | NER |
| BC5CDR [9] | 1,500 abstracts | Chemical (15,935), Disease (12,850) | NER, NEL |
| LINNAEUS [24] | 100 PMC full text | Species (4,259) | NER |
| tmVar [20] | 500 abstracts | Variant (1,431) | NER, NEL |
| NLM-Gene [14] | 550 abstracts | Gene (15,553) | NER, NEL |
| GNormPlus [28] | 694 abstracts | Gene (9,986) | NER, NEL |

Due to the limitation of the entity type in NER datasets, most of the state-of-the-art entity taggers were developed individually for a specific concept. A few studies (e.g., PubTator [29]) integrate multiple entity taggers and apply them to specific collections or even to the entire PubMed/PMC. In the development process, some challenging issues related to integrating entities from multiple taggers, such as concept ambiguity and variation emerged [30]. Moreover, the same articles need to be processed multiple times by multiple taggers. A huge storage space also is required to store the results of the taggers. In addition, based on clues from previous NER studies [28, 31], we realized that a tagger that trained with other concepts performs as well or even better than a tagger trained on only a single concept, especially for highly ambiguous concepts. A gene tagger GNormPlus trained on multiple relevant concepts (gene/family/domain) boosts the performance of a gene/protein significantly. Therefore, a rich NER corpus can help develop a method that can recognize multiple entities simultaneously to reduce the hardware requirement and achieve better performance. Only a very few datasets  [5, 27] curate multiple concepts in the text, but no relation is curated in these datasets.

### 2.2 Relation extraction

A variety of RE datasets in the general domain have been constructed to promote the development of RE systems [32-34]. Many of the RE datasets focus on extracting relations from a single sentence. Since many relations cross sentence boundaries, moving research from the sentence level to the document level (e.g., DocRED [35], DocOIE [36]) became a popular trend recently. In the biomedical domain, most existing RE datasets [6, 8, 10] focus on sentence-level relations involving a single pair of entities. However, multiple sentences are often required to describe an entire biological process or a relation. We highlight several commonly used biomedical RE datasets in Table 2 (a complete dataset review can be found in Supplementary Materials Table S6).

But only very few datasets contain relations across multiple sentences (e.g. BC5CDR dataset [9]). Most of the datasets [6-10, 37-41], which were widely used for the RE system development [42-46], focus on the single entity pair only (e.g., AIMed [38] to protein-protein interaction). Some of those datasets annotated the relation categories more granular. For example, DDI13 [8] annotated four categories (i.e., advise, int, effect, and mechanism) of the drug-drug interaction, ChemProt [10] annotated five categories of the chemical-protein interaction, and DrugProt [47], an extension of ChemProt, annotated thirteen categories. Recently, ChemProt and DDI13 are widely used in evaluating the abilities of biomedical pre-trained language models [48-51] on RE tasks.

**Table 2.** A summary of biomedical RE and event extraction datasets. The value of '-' means that we could not find the number in their papers or websites. The SEN/DOC Level means whether the relation annotation is annotated in "Sentence," "Document," or "Cross-sentence." "Document" includes abstract, full-text, or discharge record. "Cross-sentence" allows two entities within a relation to appear in three surrounding sentences.

| Dataset | # Doc./Sent. | # Entity | # Relation | SEN/DOC Level | Description |
|---|---|---|---|---|---|
| Protein-protein interaction | | | | | |
| AIMed [38] | 230 abstracts | 4,141 genes | 1,101 relations | Sentence | The AImed dataset aims to develop and evaluate protein name recognition and protein-protein interaction (PPI) extraction. It contains 750 Medline abstracts, which contain the "human" word, and has 5,206 names. Two hundred abstracts previously known to contain protein interactions for PPI extraction were obtained from the Database of Interacting Proteins (DIP) [52] and tagged for both 1,101 protein interactions and 4,141 protein names. Because negative examples for protein interactions were rare in the 200 abstracts, they manually selected 30 additional abstracts with more than one gene but did not have any gene interactions. |
| BioInfer [6] | 1,100 sentences | 4,573 proteins | 2,662 relations | Sentence | A PPI dataset uses ontologies defining the fine-granted types of entities (like "protein family or group" and "protein complex") and their relationships (like "CONTAIN" and "CAUSE"). They developed a corpus of 1100 sentences containing full dependency annotation, dependency types, and comprehensive annotation of bio-entities and their relationships. |
| BioCreative II PPI IPS [7] | 1,098 full-texts | - | - | Document | The BioCreative II PPI protein interaction pairs subtask (IPS) provides 750 and 356 full texts for training and test sets, respectively. The full-text includes corresponding gene mention symbols and PPI pairs. |
| Chemical-protein interaction | | | | | |
| DrugProt [47] | 5,000 abstracts | 65,561 chemicals, 61,775 genes | 24,526 relations | Sentence | The DrugProt dataset aims to promote the development of chemical-gene RE systems, an extension of the ChemProt dataset. It addresses 13 different chemical-gene relations, including regulatory, specific, and metabolic relations |
| Chemical-disease interaction | | | | | |
| BC5CDR [9] | 1,500 abstracts | 15,935 chemicals; 12,850 diseases | 3,106 relations | Document | BC5CDR consists of 1,500 abstracts that chemical and disease mention annotations and their IDs. It annotates chemical-induced disease relation ID pair. There are 1,400 abstracts selected from a CTD-Pfizer collaboration-related dataset, and the remaining 100 articles are new curation and are used in the test set. |
| Drug-drug interaction and Drug-ADE(adverse drug effect) interaction | | | | | |
| ADE [53] | 2,972 MEDLINE case report | 5,063 drugs; 5,776 adverse effects; 231 dosages | 6,821 drug-adverse effects; | Sentence | The ADE dataset contains drugs and conditions. But the entities do not link to the standard database identifiers. Like most of the relation datasets, ADE annotates the relations (i.e., drug-ADE and drug-dosage |

| | | | | |
|---|---|---|---|---|
| | | | 279 drug-dosage relations | relations) in sentence-level. |
| DDI13 [8] | 905 documents | 13,107 drugs | 5,028 relations | Sentence | SemEval 2013 DDIExtraction dataset consists of 792 texts selected from the DrugBank database and 233 Medline abstracts. The corpus is annotated with 18,502 pharmacological substances and 5,028 DDIs, including both pharmacokinetic (PK) and pharmaco-dynamic (PD) interactions. |
| n2c2 2018 ADE [54] | 505 summaries | 83,869 entities | 59,810 relations | - | The discharge summaries are from the clinical care database of the MIMIC-III (Medical Information Mart for Intensive Care-III). The summaries are manually selected to contain at least 1 ADE and annotated with nine concepts and eight relation pairs. The data are split into 303 and 202 for training and test sets, respectively. |

**Variant/gene-disease interaction**

| | | | | |
|---|---|---|---|---|
| EMU [21] | 110 abstracts | - | 179 relations | Document | The EMU dataset focuses on finding relationships between mutations and their corresponding disease phenotypes. They use 'MeSH = mutation' to select abstracts and use MetaMap [55] to annotate the abstracts that are divided into containing mutations related to prostate cancer (PCa) and breast cancer (BCa). They then use rules and patterns to select subsets of PCa and BCa for annotating. |
| RENET2 [56] | 1,000 abstracts, 500 full-texts | - | - | Document | It contains both 1000 abstracts (from RENET [57]) and 500 full-texts from PMC open-access subset. For better quality, 500 abstracts of the dataset were refined. The authors used the 500 abstracts to train the RENET2 model and conduct their training data expansion using the other 500 abstracts. They further used the model trained on 1,000 abstracts to construct 500 full-text articles. |

**Drug-gene-mutation**

| | | | | |
|---|---|---|---|---|
| N-ary [58] | - | - | 3,462 triples; 137,469 drug-gene relations; 3,192 drug-mutation relations; | Cross-sentence | Authors use distant supervision to construct a cross-sentence drug-gene-mutation RE dataset. They use 59 distinct drug-gene-mutation triples from the knowledge bases to extract 3,462 ternary positive relation triples. The negative instances are generated by randomly sampling the entity pairs/triples without interaction. |

**Event extraction**

| | | | | |
|---|---|---|---|---|
| GE09 [59] | 1,200 abstracts | - | 13,623 events | Sentence | As the first BioNLP shared task (ST), it aimed to define a bounded, well-defined GENIA event extraction (GE) task, considering both the actual needs and the state of the art in bio-TM technology and to pursue it as a community-wide effort. |
| GE11 [60] | 1,210 abstracts, 14 full-text | 21,616 proteins | 18.047 events | Sentence | The BioNLP ST 2011 GE task follows the task definition of the BioNLP ST 2009, which is briefly described in this section. BioNLP ST 2011 took the role of measuring the progress of the community and generalization IE technology to the full papers. |
| CG [61] | 600 abstracts | 21,683 entities | 17,248 events; 917 relations | Sentence | The BioNLP ST 2013 Cancer Genetics (CG) corpus contains annotations of over 17,000 events in 600 documents. The task addresses entities and events at all levels of biological organization, from the molecular to the whole organism, and involves pathological and physiological processes. |

During the curation of the relations in sentence-level, curators usually do not access the context of the surrounding sentences. Besides, most sentence-level RE datasets do not link the entity names to the concept identifiers (e.g., NCBI Gene ID) in the external resources/databases. Instead, the RE dataset development in document-level is highly relying on the concept identifiers. But it is extremely time-consuming, and very limited biomedical datasets annotate the relation entities to the concept identifiers. BC5CDR dataset [9] is a widely-used dataset with chemical-induced disease relations in document-level. All of the chemicals and diseases are linked to the concept identifiers. However, BC5CDR didn't annotate the relations (e.g., treatment) out of the chemical-induced disease category. Peng et al. [58] developed a cross-sentence n-ary relation extraction dataset with the relations among drug, gene, and mutation. But the dataset is constructed via distant supervision with the inevitable wrong labeling problem [35] instead of manual curation. Moreover, BioNLP shared task datasets [61-64] provide fine-grained biological event annotations to promote biological activity extraction. In Table 3, we compare BioRED to representative biomedical relation extraction datasets. BioRED covers more types of entity pairs than those datasets.

**Table 3.** Comparison of the BioRED corpus with representative relation extraction datasets. D = Disease, G = Gene, C = Chemical, and V = Variant.

| | <D,G> | <D,C> | <D,V> | <C,C> | <C,G> | <G,G> | <V,C> | <V,V> |
|---|---|---|---|---|---|---|---|---|
| RENET2 | ✓ | | | | | | | |
| BC5CDR | | ✓ | | | | | | |
| EMU | | | ✓ | | | | | |
| DDI13 | | | | ✓ | | | | |
| DrugProt | | | | | ✓ | | | |
| AIMed | | | | | | ✓ | | |
| GE11 | | | | | | ✓ | | |
| N-ary | | | | | ✓ | | ✓ | |
| CG | ✓ | ✓ | | ✓ | ✓ | ✓ | | |
| BioRED | ✓ | ✓ | ✓ | ✓ | ✓ | ✓ | ✓ | ✓ |

## 3 Methods

### 3.1 Annotation definition/scope

We first analyzed a set of public PubMed search queries by tagging different entities and relations. This data-driven approach allowed us to determine a set of key entities and relations of interest that should be most representative, and therefore the focus of this work. Some entities are closely related biologically and are thus used interchangeably in this work. For instance, protein, mRNA, and some other gene products typically share the same names and symbols. Thus, we merged them to a single gene class, and similarly merged symptoms and syndromes to a single disease class. In the end, we have six concept types: (1) Gene: for genes, proteins, mRNA and other gene products. (2) Chemical: for chemicals and drugs, (3) Disease: for diseases, symptoms, and some disease-related phenotypes. (4) Variant: for genomic/protein variants (including substitutions, deletions, insertions, and others). (5) Species: for species in the hierarchical taxonomy of organisms. (6) CellLine: for cell lines. Due to the critical problems of term variation and ambiguity, entity linking (also called entity normalization) is also required. We linked the entity spans to specific identifiers in an appropriate database or controlled vocabulary for each entity type (e.g., NCBI Gene ID for genes).

Between any of two different entity types, we further observed eight popular associations that are frequently discussed in the literature: <D,G> for <Disease, Gene>; <D,C> for <Disease, Chemical>, <G,C> for <Gene,

Chemical>, <G,G> for <Gene, Gene>, <D,V> for <Disease, Variant>, <C,V> for <Chemical, Variant>, <C,C> for <Chemical, Chemical> and <V,V> for <Variant, Variant>. For relations between more than two entities, we simplified the relation to multiple relation pairs. For example, we simplified the chemicals co-treat disease relation ("bortezomib and dexamethasone co-treat multiple myeloma") to three relations: <bortezomib, multiple myeloma, treatment>, <dexamethasone, multiple myeloma, treatment>, and <bortezomib, dexamethasone, co-treatment> (treatment is categorized in the Negative_Correlation). Other associations between two concepts are either implicit (e.g., variants frequently located within a gene) or rarely discussed. Accordingly, in this work we focus on annotating those eight concept pairs, as shown in solid lines in Figure 2a. To further characterize relations between entity pairs, we used eight biologically meaningful and non-directional relation types (e.g., positive correlation; negative correlation) in our corpus as shown in Figure 2b. The details of the relation types are described in our annotation guideline.

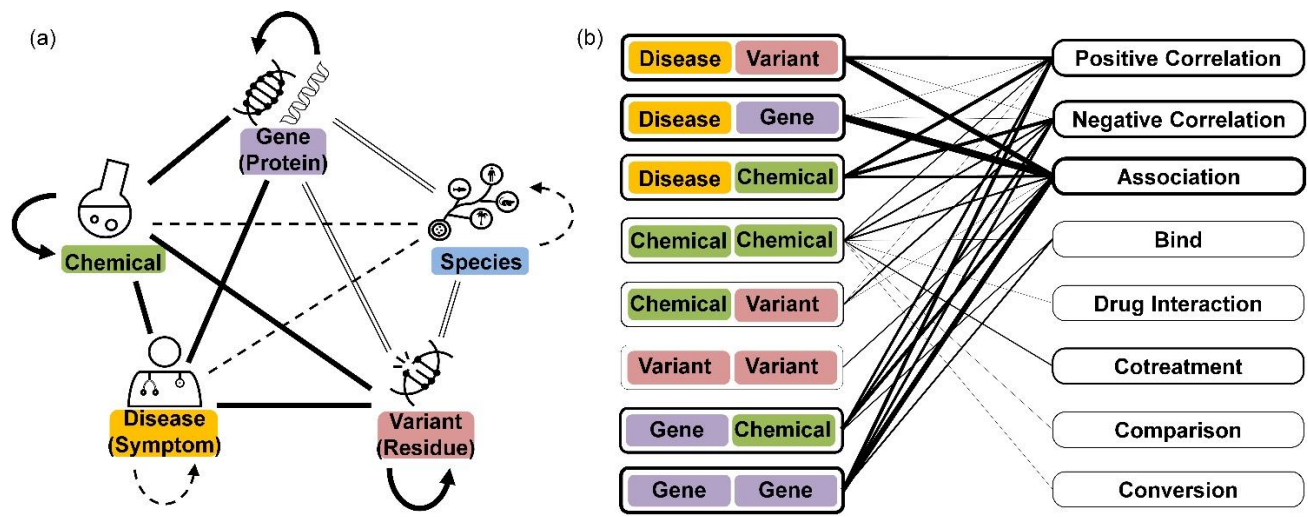

**Figure 2.** Relations annotated in BioRED corpus. (a) Categorized relations between concepts. The patterns of the lines between the concepts present the categories: (━) Popular associations: The concept pairs are frequently discussed in the biomedical literature. (═) Implied associations e.g., the name of a gene can imply the corresponding species. (---) Rarely discussed associations: Some other relation types are rarely discussed in the biomedical text (and this is why the concept Cell Line is not listed here). (b) The mapping between the concept pairs and the relation types. The frame widths of the concept pairs/relation types and the bold lines between the two sides proportionally represent the frequencies

## 3.2 Annotation process

In order to be consistent with previous annotation efforts, we randomly sampled articles from several existing datasets (i.e., NCBI Disease[19], NLM-Gene [14], GNormPlus [28], BC5CDR [9], tmVar [20, 62]). A small set of PubMed articles were first used to develop our annotation guidelines and familiarize our annotators with both the task and TeamTat [63], a web-based annotation tool equipped to manage team annotation projects efficiently. Following previous practice in biomedical corpus development, we developed our annotation guidelines and selected PubMed articles consistently with previous studies. Furthermore, to accelerate entity annotation, we used previous annotations combined with automated pre-annotations (i.e., PubTator [29]), which can then be edited based on human judgment. Unlike entity annotation, each relation is annotated from scratch by hand with an appropriate relation type, except the chemical-induced-disease relations that were previously annotated in BC5CDR.

Every article in the corpus was first annotated by three annotators with background in biomedical informatics to prevent erroneous and incomplete annotations (especially relations) due to manual annotation fatigue. If an entity or a relation cannot be agreed upon by the three annotators, this annotation was then reviewed by another senior annotator with background in molecular biology. For each relation, two additional biologists assessed whether it is novel finding vs. background information and made the annotation accordingly. We annotated the entire set of 600 abstracts in 30 batches of 20 articles each. For each batch, it takes approximately 2 hours per annotator to annotate entities, 8 hours for relations, and 6 hours for assigning novel vs. background label. The details of the data sampling and annotation rules are described in our annotation guideline.

### 3.3 Data Characteristics

The BioRED corpus contains a total of 20,419 entity mentions, corresponding to 3,869 unique concept identifiers. We annotated 6,503 relations in total. The proportion of novel relations among all annotated relations in the corpus is 69%. Table 4 shows the numbers of the entities (mentions and identifiers) and relations in the training, development, and test sets.

**Table 4.** Number of entity (mention and identifier) and relation annotations in the BioRED corpus, the inter-annotator-agreement (IAA), and the distribution between the training, development, and test sets. The parenthesized numbers are the unique entities linked with concept identifiers.

| Annotation | | Training | Dev | Test | Total | IAA |
|---|---|---|---|---|---|---|
| Document | | 400 | 100 | 100 | 600 | - |
| Entity (ID) | All | 13,351 (2,708) | 3,533 (956) | 3,535 (982) | 20,419 (3,869) | 97.01% |
| | Gene | 4,430 (1,141) | 1,087 (368) | 1,180 (399) | 6,697 (1,643) | 97.35% |
| | Disease | 3,646 (576) | 982 (244) | 917 (244) | 5,545 (778) | 96.06% |
| | Chemical | 2,853 (486) | 822 (184) | 754 (170) | 4,429 (651) | 96.12% |
| | Variant | 890 (420) | 250 (135) | 241 (137) | 1,381 (678) | 97.79% |
| | Species | 1,429 (37) | 370 (13) | 393 (11) | 2,192 (47) | 99.43% |
| | Cell Line | 103 (48) | 22 (12) | 50 (21) | 175 (72) | 99.68% |
| Relation | | 4,178 | 1,162 | 1,163 | 6,503 | 77.91% |
| Relation pair with novelty findings | | 2,838 | 835 | 859 | 4,532 | 85.01% |

In addition, we computed the inter-annotator-agreement (IAA) for entity, relation, and novelty annotations, where we achieved 97.01%, 77.91%, and 85.01%, respectively. Figure 3 depicts the distribution of the different concept pairs in the relations.

We also analyzed dataset statistics per document. The average document length consists of 11.9 sentences or 304 tokens. 34 entity spans (3.8 unique entity identifiers) and 10.8 relations are annotated per document. Among the relation types, 52% are associations, 27% are positive correlations, 17% are negative correlations, and 2% are involved in the triple relations (e.g., two chemicals co-treat a disease).

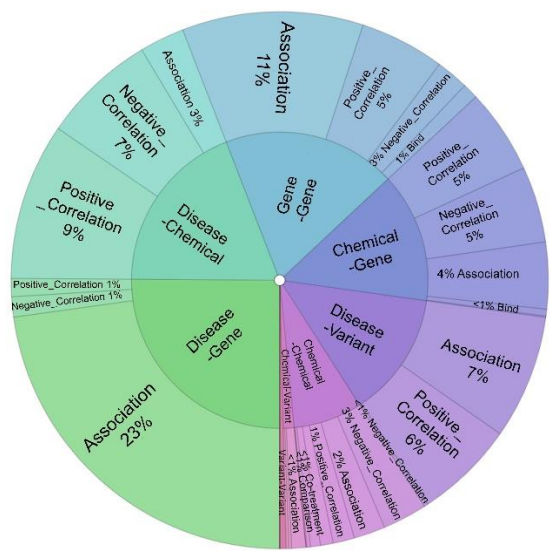

**Figure 3.** The distribution of concept pairs and relation types in the BioRED corpus.

## 3.4 Benchmarking methods

To assess the utility and challenges of the BioRED corpus, we conducted experiments to show the performance for leading RE models. For the NER task, each mention span was considered separately. We evaluate three state-of-the-art NER models on the corpus including BiLSTM-CRF, BioBERT-CRF and PubMedBERT-CRF. The input documents are first split into multiple sentences and encoded into a hidden state vector sequence by Bidirectional Long Short-Term Memory (BiLSTM) [64], BioBERT [51], PubMedBERT [49], respectively. The models predicted the label corresponding to each of the input tokens in the sequence, and then computed the network score using a fully connected layer, and decode the best path of the tags in all possible paths by using Conditional Random Field (CRF) [65]. Here, we used the BIO (Begin, Inside, Outside) tagging scheme to the CRF layer.

We chose two BERT-based models, BERT-GT [66] and PubMedBERT [67], for evaluating the performance of current RE systems on the BioRED corpus. The first model is BERT-GT, which defines a graph transformer through integrating a neighbor–attention mechanism into the BERT architecture to avoid the effect of the noise from the longer text. BERT-GT was specifically designed for document-level relation extraction tasks and utilizes the entire sentence or passage to calculate the attention of the current token, which brings significant improvement to the original BERT model. PubMedBERT is a pretrained biomedical language model based on transformer architecture. It is currently a state-of-the-art text-mining method, which applies the biomedical domain knowledge (biomedical text and vocabulary) for the BERT pretrained language model. In the benchmarking, we used the text classification framework for the RE model development.

For both NER and RE evaluations, the training and development sets were first used for model development and parameter optimization before a trained model is evaluated on the test set. Benchmark implementation details are provided in Supplementary Materials A.1. Standard Precision, Recall and F-score metrics are used. To allow approximate entity matching, we also applied relaxed versions of F-score to evaluate NER. In this case, as long as the boundary of the predicted entity overlaps with the gold standard span, it is considered as a successful prediction.

## 4   Results

## 4.1 NER results on the test set

Table 5 shows the evaluation of NER on the test set. The first run is evaluated by strict metrics. The concept type and boundary of the entity should exactly match the entity in the text. The second run is evaluated by relaxed metrics: The entity boundary should overlap, and the same entity type is required. Unlike BiLSTM-CRF, the BERT-based methods contain well pre-trained language models for extracting richer features, hence achieving better performance overall. Further, PubMedBERT performs even better than BioBERT on genes, variants, and cell lines. BioBERT uses the original BERT model's vocabulary generated from general domain text, which causes the lack of understanding on the biomedical entities. On the contrary, PubMedBERT generates the vocabulary from scratch using biomedical text, and it achieves the highest F-score (89.3% in strict metric). Among these entity types, the PubMedBERT-CRF achieves the highest performance of 97% in F1-score to species entity recognition as less term ambiguity and variation issues are fond in species names.

**Table 5.** Performance of NER models on test set. All numbers are F-scores. G = Gene, D = Disease, C = Chemical, S = Species, CL = CellLine, and V = Variant.

| Metric | Method | All | G | D | C | S | CL | V |
|--------|--------|-----|-----|-----|-----|-----|-----|-----|
| Strict | BiLSTM-CRF | 87.1 | 87.3 | 83.3 | 88.2 | 96.3 | 80.9 | 82.9 |
| | BioBERT-CRF | 88.7 | 89.5 | **84.8** | **89.7** | 96.7 | 83.5 | 83.9 |
| | PubMedBERT-CRF | **89.3** | **92.4** | 83.5 | 88.6 | **97.0** | **90.5** | **87.3** |
| Relaxed | BiLSTM-CRF | 92.4 | 92.3 | 92.2 | **91.9** | 96.8 | 85.4 | 93.6 |
| | BioBERT-CRF | 93.4 | 93.8 | **93.6** | 91.3 | **97.0** | 90.1 | 92.3 |
| | PubMedBERT-CRF | **93.5** | **94.7** | 92.6 | 91.1 | **97.0** | **92.6** | **94.5** |

## 4.2 RE results on the test set

We also evaluated performance on the RE task by different benchmark schemas: (1) entity pair: to extract the pair of concept identifiers within the relation, and (2) entity pair + relation type: to recognize the specific relation type for the extracted pairs, and (3) entity pair + relation type + novelty: to further label the novelty for the extracted pairs. In this task, the gold-standard concepts in the articles are given. We applied BERT-GT and PubMedBERT to recognize the relations and the novelty in the test set.

As shown in Table 6, the overall performance of PubMedBERT is higher than that of BERT-GT in all schemas. Because the numbers of relations in <D,V>, <C,V> and <V,V> are low, their performance is not comparable to that of other concept pairs, especially <V,V> (the F-score is 0% for two models). In the first schema, BERT-GT and PubMedBERT can achieve performance above 72% for the F-scores, which is expected and promising in the document-level RE task. To predict the relation types (e.g., positive correlation) other than entity pairs, however, is still quite challenging. The best performance on the second schema is only 58.9%, as the number of instances in many relation types is insufficient. The performances on different relation types of our best model using PubMedBert are provided in Supplementary Materials A.2. The performance on the third schema dropped to 47.7%. In some cases, the statements of the relations in abstracts are usually concise, and the details of the relation mechanism can only be found in the full text.

**Table 6.** Performance on RE task for the first schema: extracting the entity pairs within a relation, second schema: extracting the entity pairs and the relation type, and the third schema: further labeling the novelty for the extracted pairs. All numbers are F-scores. The <G,D> is the concept pair of the gene (G) and the disease

(D). The columns of those entity pairs present the RE performance in F-scores. G = gene, D = disease, V = variant, and C = chemical.

| Eval Schema | Method | All | <G,D> | <G,G> | <G,C> | <D,V> | <C,D> | <C,V> | <C,C> |
|---|---|---|---|---|---|---|---|---|---|
| Entity pair | BERT-GT | 72.1 | 63.8 | **78.5** | 77.7 | **69.8** | 76.2 | **58.8** | 74.9 |
| | PubMedBERT | **72.9** | **67.2** | 78.1 | **78.3** | 67.9 | **76.5** | 58.1 | **78.0** |
| +Relation type | BERT-GT | 56.5 | 54.8 | 63.5 | **60.2** | 42.5 | **67.0** | 11.8 | 52.9 |
| | PubMedBERT | **58.9** | **56.6** | **66.4** | 59.9 | **50.8** | 65.8 | **25.8** | **54.4** |
| +Novelty | BERT-GT | 44.5 | 37.5 | 47.3 | **55.0** | 36.9 | **51.9** | 11.8 | 48.5 |
| | PubMedBERT | **47.7** | **40.6** | **54.7** | 54.8 | **42.8** | 51.6 | **12.9** | 50.3 |

## 4.3 Benefits of multiple entity recognition and relation extraction.

To test the hypothesis that our corpus can result in a single model with better performance, we trained multiple separate NER and RE models, each with an individual concept (e.g., gene) or relation (e.g., gene-gene) for comparison. We used PubMedBERT for this evaluation since it achieved the best performances in both the NER and RE tasks. As shown in Table 7, both models trained on all entities or relations generally perform better than the models trained on most of the entities or relations, while the improvement for RE is generally larger. The performance on NER and RE tasks are both obviously higher in the single model. Especially for entities and relations (e.g., cell lines and chemical-chemical relations) with insufficient amounts, the model trained on multiple concepts/relations can obtain larger improvements. The experiment demonstrated that training NER/RE models with more relevant concepts or relations not only can reduce resource usage but also can achieve better performance.

**Table 7.** The comparison of the models trained on all entities/relations to the models trained on individual entity/relation. The <G,D> is the relation of the gene (G) and the disease (D). G = gene, D = disease, C = chemical, V = variant, S = species, and CL = cell line. All models are evaluated by strict metrics.

| Entity/Relation | Type | All entities or relations | | | Single entity or relation | | |
|---|---|---|---|---|---|---|---|
| | | P | R | F | P | R | F |
| Entity | G | 92.2 | 92.5 | **92.4** | 90.8 | 91.0 | 90.9 |
| | D | 80.7 | 86.5 | 83.5 | 83.2 | 85.7 | **84.4** |
| | C | 87.9 | 89.3 | 88.6 | 87.3 | 92.4 | **89.8** |
| | V | 88.8 | 85.9 | **87.3** | 84.7 | 87.1 | 85.9 |
| | S | 95.8 | 98.2 | **97.0** | 95.2 | 96.4 | 95.8 |
| | CL | 95.6 | 86.0 | **90.5** | 77.1 | 74.0 | 75.5 |
| Relation | <G,D> | 63.6 | 71.2 | 67.2 | 75.8 | 62.7 | **68.7** |
| | <G,G> | 81.5 | 75.0 | **78.1** | 57.3 | 80.0 | 66.8 |
| | <G,C> | 74.1 | 83.1 | **78.3** | 66.7 | 68.9 | 67.8 |
| | <D,V> | 71.2 | 64.9 | **67.9** | 76.5 | 51.5 | 61.5 |
| | <C,D> | 73.3 | 79.9 | 76.5 | 78.2 | 85.2 | **81.5** |
| | <C,V> | 60.0 | 56.3 | **58.1** | 53.3 | 50.0 | 51.6 |
| | <C,C> | 75.3 | 80.9 | **78.0** | 64.2 | 72.3 | 68.0 |

## 4.4 Discussion

The relaxed NER results in Table 5 for overall entity type are over 92% for all methods, suggesting the maturity of current tools for this task. If considering the performance of each concept individually, the recognition of genes, species and cell lines can reach higher performance (over 90% in strict F-score) since the names are often simpler and less ambiguous than other concepts. The best model for genomic variants achieves an F-score of 87.3% in strict metrics and 94.5% in relaxed metrics, which suggests that the majority of the errors are due to incorrect span boundaries. Most variants are not described in accordance with standard nomenclature (e.g., "ACG-->AAG substitution in codon 420"), thus it is difficult to exactly identify the boundaries. Like genomic variants, diseases are difficult to be identified due to term variability and most errors are caused by mismatched boundaries. For example, our method recognized a part ("papilledema") of a disease mention ("bilateral papilledema") in the text. Disease names also present greater diversity than other concepts: 55.4% of the disease names in the test set are not present in the training/development sets. Chemical names are extremely ambiguous with other concepts: half of the errors for chemicals are incorrectly labeled as other concept types (e.g., gene), since some chemicals are interchangeable with other concepts, like proteins and drugs. Moreover, we merged the annotations matched by the dictionary to the results of the PubMedBERT-CRF model. However, the performance of the dictionary method heavily depends on the difficulties of the term variation and ambiguity issues. Especially, there are many ambiguous terms in dictionary, such like "B1", "Beta" and "98-4.9" in Cellosaurus. Although the F1-score of the dictionary cannot compete with the machine learning method, merging the results from both methods can improve the recall for all the concepts (see details in Supplementary Materials A.3).

Experimental results in Table 6 show that the RE task remains challenging in biomedicine, especially for the new task of extracting novel findings. In our observation, there are three types of errors in novelty identification. First, some abstracts do not indicate which concept pairs represent novel findings, and instead provide more details in the full text. Such cases confused both the human annotators and the computer algorithms. Second, when the mechanism of interaction between two relevant entities is unknown, and the study aims to investigate it but the hypothesized mechanism is shown to be false. Third, the authors frequently mention relevant background knowledge within their conclusion. As an example, "We conclude that Rg1 may significantly improve the spatial learning capacity impaired by chronic morphine administration and restore the morphine-inhibited LTP. This effect is NMDA receptor dependent." in the conclusion of the PMID:18308784, the Rg1 responded to morphine as a background knowledge. But it is mentioned together with the novelty knowledge pair <Rg1, NMDA receptor>. In this case, our method misclassified the pair < Rg1, morphine> as Novel. We also conducted an experiment to evaluate the effect of section information for novelty detection. The experimental results show that the structured section information (e.g., TITLE, PURPOSE, METHODS, RESULTS, ...) can be useful for novelty classification by boosting the best F1-score from 47.7% to 48.9% (see details in Supplementary Materials A.4). However, this result was obtained on a subset of 191 abstracts with structured section information due to limited availability.

The results in Table 7 demonstrate that training NER/RE models on one rich dataset with multiple concept/relations simultaneously can not only make the trained model simpler and more efficient, but also more accurate. More importantly, we notice that for the entities and relations with a lower number of training instances (e.g., cell lines and chemical-chemical relations), simultaneous prediction is especially beneficial for improving performance. Additionally, merging entity results from different models often poses some challenges, such as ambiguity or overlapping boundaries between different concepts.

## 5   Conclusion

In the past, biomedical RE datasets were typically built for a single entity type or relation. To enable the development of RE tools that can accurately recognize multiple concepts and their relations in biomedical texts, we have developed BioRED, a high-quality RE corpus, with one-of-a-kind novelty annotations. Like other commonly used biomedical datasets, e.g., BC5CDR [9], we expect BioRED to serve as a benchmark for not only biomedical-specific NLP tools but also for the development of RE methods in general domain. Additionally, the novelty annotation in BioRED proposes a new NLP task that is critical for information extraction in practical applications. Recently, the dataset was successfully used by the NIH LitCoin NLP Challenge (https://ncats.nih.gov/funding/challenges/litcoin) and a total of over 200 teams participated in the Challenge.

This work has implications for several real-world use cases in medical information retrieval, data curation, and knowledge discovery. Semantic search has been commonly practiced in the general domain but much less so in biomedicine. For instance, several existing studies retrieve articles based on the co-occurrence of two entities [68-71] or rank search results by co-occurrence frequency. Our work could accelerate the development of semantic search engine in medicine. Based on the extracted relations within documents, search engines can semantically identify articles by two entities with relations (e.g., 5-FU-induced cardiotoxicity) or by expanding the user queries from an entity (e.g., 5-FU) to the combination of the entity and other relevant entities (e.g., cardiotoxicity, diarrhea).

While BioRED is a novel and high-quality dataset, it has a few limitations. First, we are only able to include 600 abstracts in the BioRED corpus due to the prohibitive cost in manual annotation and limited resources. Nonetheless, our experiments show that except for few concept pairs and relation types (e.g. variant-variant relations) that occur infrequently in the literature, its current size is appropriate for building RE models. Our experimental results in Table 7 also show that in some cases, the performance on entity class with a small number of training instances (e.g. Cell Line) can be significantly boosted when training together with other entities in one corpus. Second, the current corpus is developed on PubMed abstracts, as opposed to full text. While full text contains more information, data access remains challenging in real-world settings. More investigation is warranted on this topic in the future.

## Acknowledgements

The authors are grateful to Drs. Tyler F. Beck and Christine Colvis, Scientific Program Officer at the NCATS and their entire research team for help with our dataset. The authors would like to thank Rancho BioSciences and specifically, Mica Smith, Thomas Allen Ford-Hutchinson, and Brad Farrell for their contribution with data curation.

## Funding

This work was supported by the National Institutes of Health intramural research program, National Library of Medicine and partially supported by the NIH grant 2U24HG007822-08 to CNA.
*Conflict of Interest: none declared.*

# BioRED: A Rich Biomedical Relation Extraction Dataset
## (Supplementary Materials)

## A.1 Benchmark implementation details

Here we provide the implementation details of our methods. We firstly selected the hyper-parameters by random search [1] on the development set. Then we merged the training and development sets to retrain the model. The number of training epochs is determined by the early stopping strategy [2] according to the training loss. All models were trained and tested on the NVIDIA Tesla V100 GPU.

NER models: We evaluate three state-of-the-art NER models including BiLSTM-CRF, BioBERT-CRF and PubMedBERT-CRF. We used concatenation of word embedding and character-level features generated with a CNN input layer for BiLSTM-CRF. The two BERT-based models used BioBERT-Base-Cased v1.1[1] and PubMedBERT-base-uncased-abstract[2] with default parameter settings to build the encoders via the Hugging Face platform. We optimized BiLSTM-CRF using RMSprop with a learning rate of 1e-3 The BERT-based models used Adam with a learning rate of 1e-5. The other experimental hyper-parameters are shown in Table S1.

**Table S1**. NER Hyper-parameter settings

| General Hyper-parameter | |
| --- | --- |
| Batch size | 32 |
| Epochs at most | 50 |
| Fully connection size | 128 |
| BiLSTM-CRF Hyper-parameter | |
| Character-level CNN hidden size | 100 |
| Character-level CNN window size | 3 |
| Word-level LSTM hidden size | 512 |
| Word-level LSTM dropout rate | 0.4 |
| Word embedding dimension | 200 |
| Character embedding dimension | 50 |

RE models: We applied two state-of-the-art RE models, PubMedBert and BERT-GT for both RE and novelty triage tasks. We first use two tags [SourceEntity] and [TargetEntity] to represent the source entities and target entities. Then, the tagged abstract turns to a text sequence as the input of the models. We use the [CLS]'s hidden layer and a softmax layer in the classification. We applied the source codes provided by BERT-GT to convert the corpus. BERT-GT used the pre-trained language model of BioBERT. The detailed hyper-parameters of both tasks are shown in Table S2.

**Table S2**. Hyper-parameter settings for RE and Novelty triage.

| | RE | | Novelty | |
| --- | --- | --- | --- | --- |
| | PubMedBERT | Bert-GT | PubMedBERT | Bert-GT |
| batch size | 16 | 8 | 16 | 8 |
| epochs | 10 | 30 | 10 | 30 |
| learning rate | 1e-5 | 1e-5 | 1e-5 | 1e-5 |
| sequence length | 512 | 512 | 512 | 512 |
| the others | default | default | default | default |

---

[1] https://huggingface.co/dmis-lab/biobert-base-cased-v1.1

[2] https://huggingface.co/microsoft/BiomedNLP-PubMedBERT-base-uncased-abstract

## A.2 Performances of different relation types on the test set

Here, we detailed the performances on different relation types of our best model using PubMedBert on the test set. The results are shown in Table S3. We filled "-" in the table if the relation type doesn't exist in the entity pairs.

**Table S3**. Performance of different relation types on relation extraction (RE) task. All numbers are F-scores. The <G,D> is the concept pair of the gene (G) and the disease (D). G = gene, D = disease, V = variant, and C = chemical.

| Relation Type | <G,D> | <G,G> | <G,C> | <D,V> | <C,D> | <C,V> | <C,C> |
|---|---|---|---|---|---|---|---|
| Association | 60.0 | 61.9 | 45.6 | 51.5 | 25.5 | 32.6 | 25.5 |
| Positive_Correlation | 7.7 | 79.1 | 61.9 | 50.0 | 76.6 | 0.0 | 47.8 |
| Negative_Correlation | 30.8 | 54.1 | 79.1 | 0.0 | 61.9 | 0.0 | 76.1 |
| Cotreatment | - | - | 66.7 | - | - | - | 60.0 |
| Drug_Interaction | - | - | 0.0 | - | - | - | 66.7 |
| Bind | - | 57.1 | 54.5 | - | - | - | - |
| Comparison | - | - | - | - | - | - | 50.0 |
| Conversion | - | - | - | - | - | - | 0.0 |
| Overall | 56.6 | 66.4 | 59.9 | 50.8 | 65.8 | 25.8 | 54.4 |

## A.3 Performances of dictionary-based method on the test set

Moreover, we also implemented a dictionary-based method to complement the pre-trained model. We used the term names and synonyms in the latest version of CTD-chemical (http://ctdbase.org/reports/CTD_chemicals.tsv.gz), CTD-disease (http://ctdbase.org/reports/CTD_diseases.tsv.gz) and Cellosaurus (https://ftp.expasy.org/databases/cellosaurus/cellosaurus.obo) to construct the chemical, disease and cell line dictionaries, respectively. Then the prefix search is applied for exact dictionary matching. As the result shown in Table S4, the performance of the dictionary method heavily depends on the difficulty of the term variation and ambiguity issues. Especially, there are many ambiguous terms in the dictionary, such like "B1", "Beta" and "98-4.9" in Cellosaurus. Even though the F1-score of the dictionary can't compete with the machine learning method. But merging the results from both methods can improve the recall for all the concepts. In future work, we will further explore the way to use the additional features by dictionary-match to train the deep learning models.

**Table S4**. Performances of dictionary-based method on the test set

| Method | Strict metrics | | | Relaxed metrics | | |
|---|---|---|---|---|---|---|
| | Precision | Recall | F-score | Precision | Recall | F-score |
| Disease-Dictionary | 75.5 | 64.0 | 69.3 | 91.0 | 76.0 | 82.8 |
| Disease-PubMedBERT | 80.7 | 86.5 | 83.5 | 91.2 | 96.1 | 93.6 |
| Disease-Dictionary+PubMedBERT | 77.9 | 87.1 | 82.2 | 88.0 | 96.8 | 92.2 |
| Chemical-Dictionary | 58.7 | 78.5 | 67.2 | 61.2 | 81.3 | 69.8 |
| Chemical-PubMedBERT | 87.9 | 89.3 | 88.6 | 90.6 | 92.0 | 91.3 |
| Chemical-Dictionary+PubMedBERT | 75.6 | 89.8 | 82.1 | 77.9 | 92.8 | 84.7 |
| Cellline-Dictionary | 8.86 | 84.0 | 16.0 | 10.1 | 94.0 | 18.3 |
| Cellline-PubPedBERT | 95.6 | 86.0 | 90.5 | 97.8 | 88.0 | 92.6 |
| Cellline-Dictionary+PubMedBERT | 17.9 | 88.0 | 29.7 | 18.3 | 90.0 | 30.4 |

## A.4 The effect of structured section information for novelty detection

We also developed a model (i.e., "PubMedBERT+ Structure") using PubMedBERT to explore if the argumentative structure of the abstract (e.g., TITLE, PURPOSE, METHODS, RESULTS, ...) can help with the classification of the novelty. As we can collect from PubMed section categories (https://lhncbc.nlm.nih.gov/ii/areas/structured-abstracts/downloads.html), 191 abstracts (155 in the training set and 36 in the test set) in BioRED are with structured section. For an example of the result section in PMID: 20105280, the input sequence turns to *"… corneas . <RESULTS>RESULTS</RESULTS> : [SourceEntity] Ras transgenic lenses …. increases in [TargetEntity] cyclin D1 and D2 expression …"* ( "cyclin D1" and "Ras" are within a relation). As shown in Table S5, the performance has been increased from 47.7% to 48.9% F1-score using the section information. Due to the significant contribution of the section information, we will explore the way to automatically extract the sections of the abstracts in the future.

**Table S5**. The effect of structured section information on performance for novelty detection

| Method | All | <G,D> | <G,G> | <G,C> | <D,V> | <C,D> | <C,V> | <C,C> |
|---|---|---|---|---|---|---|---|---|
| PubMedBERT | 47.7 | 40.6 | 54.7 | 54.8 | 42.8 | 51.6 | 12.9 | 50.3 |
| PubMedBERT+Structure | 48.9 | 42.2 | 56.4 | 55.4 | 43.4 | 54.7 | 12.9 | 49.2 |

**Table S6.** Overview of biomedical RE and event extraction datasets. The value of '-' means that we could not find the number in their papers or websites. The SEN/DOC Level means whether the relation annotation is annotated in "Sentence," "Document," or "Cross-sentence." "Document" includes abstract, full-text, or discharge record. "Cross-sentence" allows two entities within a relation to appear in three surrounding sentences.

| Dataset | # Doc./Sent. | # Entity | # Relation | SEN/DOC Level | Description |
|---|---|---|---|---|---|
| Protein-protein interaction | | | | | |
| AIMed [3] | 230 abstracts | 4,141 genes | 1,101 relations | Sentence | The AImed dataset aims to develop and evaluate protein name recognition and protein-protein interaction (PPI) extraction. It contains 750 Medline abstracts, which contain the "human" word, and has 5,206 names. Two hundred abstracts previously known to contain protein interactions for PPI extraction were obtained from the Database of Interacting Proteins (DIP) [4] and tagged for both 1,101 protein interactions and 4141 protein names. Because negative examples for protein interactions were rare in the 200 abstracts, they manually selected 30 additional abstracts with more than one gene but did not have any gene interactions. |
| HPRD50 [5] | 50 abstracts | - | 138 relations | Sentence | They randomly selected 50 abstracts (called hprd50) from the Human Protein Reference Database (HPRD) [6] and manually annotated PPI, involving direct physical interactions, regulatory relations, and modifications (e.g., phosphorylation). There are 138 gene/protein relation pairs and 92 distinct pairs in abstracts. |
| BioInfer [7] | 1100 sentences | 4,573 proteins | 2,662 relations | Sentence | A PPI dataset uses ontologies defining the fine-granted types of entities (like "protein family or group" and "protein complex") and their relationships (like "CONTAIN" and "CAUSE"). They developed a corpus of 1,100 sentences containing full dependency annotation, dependency types, and comprehensive annotation of bio-entities and their relationships. |
| IEPA[8] | 300 | - | - | Document | The Interaction Extraction Performance Assessment (IEPA) |
| | abstracts | | | | corpus consists of ~300 abstracts retrieved from MEDLINE using ten queries. Each query was the AND of two biochemical nouns which domain experts suggested. The studied set included approximately forty abstracts describing interaction(s) between the biochemicals in the query, plus those that contained the biochemicals but did not describe interactions between them that were also encountered. Thus the ten queries yielded ten sets of abstracts, with each abstract in a set containing both terms in the query corresponding to that set. |
| LLL [9] | 167 sentences | - | 377 relations | Sentence | The LLL05 challenge task aims to learn rules to extract protein/gene interactions in the form of relations from biology abstracts from the Medline bibliography database. The challenge aims to test the ability of ML systems to learn rules for identifying the gene/proteins that interact and their roles, agent or target. |
| BioCreative II PPI IPS [10] | 1,098 full-texts | - | - | Document | The BioCreative II PPI protein interaction pairs subtask (IPS) provides 750 and 356 full texts for training and test sets, respectively. The full-text includes corresponding gene mention symbols and PPI pairs. |
| BioCreative II.5 IPT [11] | 122 full-texts | - | - | Document | The BioCreative II.5 interaction pair task (IPT) provide 595 full-texts for both training (FEBS Letters articles from 2008) and test (FEBS Letters articles from 2007) sets. The full-texts include both with and without curatable protein interactions, and only 122 full-texts contain PPI annotations. |
| BioCreative VI PM[12] | 5,509 abstracts | - | 1,232 relations | Document | BC6PM contains PubMed abstracts (from IntAct/Mint [13]) annotated with those interacting PPI pairs affected by mutations. The relation annotation is represented in Entrez Gene ID pair. |
| Chemical-protein interaction | | | | | |
| ChemProt [14] | 2,482 abstracts | 32,514 chemicals, 30,912 genes | 10,270 relations | Sentence | The ChemProt dataset consists of manually annotated chemical compound/drug and gene/protein mentions and 22 different chemical-protein relation types. Five relation types are used for evaluation, including agonist, antagonist, inhibitor, activator, and substrate/product relations. |
| DrugProt [15] | 5,000 abstracts | 65,561 chemicals; 61,775 genes | 24,526 relations | Sentence | The DrugProt dataset aims to promote the development of chemical-gene RE systems, an extension of the ChemProt dataset. The addressed 13 different chemical-gene relations, including regulatory, specific, and metabolic relations |
| Chemical-disease interaction | | | | | |
| BC5CDR [16] | 1,500 abstracts | 15,935 chemicals; 12,850 diseases | 3,106 | Document | CDR consists of 1,500 abstracts that chemical and disease mention annotations and their IDs. It annotates chemical-induced disease relation ID pair. There are 1,400 abstracts selected from a CTD-Pfizer collaboration-related dataset, and the remaining 100 articles are new curation and are used in the test set. |
| Drug-drug interaction and Drug-ADE interaction | | | | | |
| ADE [17] | 2,972 MEDLINE case report | 5,063 drugs; 5,776 adverse effects; 231 dosages | 6,821 drug-adverse effects; 279 drug-dosage relations | Sentence | The ADE dataset contains drugs and conditions. But the entities do not link to the standard database identifiers. Like most of the relation datasets, ADE annotate the relation (i.e., drug-ADE and drug-dosage rela-tions) in sentence-level. |
| DDI13 [18] | 905 | 13,107 drugs | 5,028 | Sentence | SemEval 2013 DDIExtraction dataset consists of 792 texts selected from the DrugBank database and 233 Medline abstracts. The corpus is annotated with 18,502 pharmacological substances and 5,028 DDIs, including both pharmacokinetic (PK) and pharmacodynamic (PD) interactions. |
| n2c2 2018 ADE[19] | 505 summaries | 83,869 entities | 59,810 relations | - | The discharge summaries are from the clinical care database of the MIMIC-III (Medical Information Mart for Intensive Care-III). The summaries are manually selected to contain at least 1 ADE and annotated with nine concepts and eight relation pairs. The data are split into 303 and 202 for training and test sets, respectively. |
| Variant/gene-disease interaction | | | | | |
| EMU [20] | 110 abstracts | - | 179 relations | Document | The EMU dataset focuses on finding relationships between mutations and their corresponding disease phenotypes. They use 'MeSH = mutation' to select abstracts and use MetaMap [21] to |

|  |  |  |  |  | annotate the abstracts that are divided into containing mutations related to prostate cancer (PCa) and breast cancer (BCa). They then use rules and patterns to select subsets of PCa and BCa for annotating. |
| RENET2 [22] | 1,000 abstracts; 500 full-text | - | - | Document | It contains both 1000 abstracts (from RENET[23]) and 500 full-texts from PMC open-access subset. For a better quality, 500 abstractsof the dataset were refined. Authors used the 500 abstracts to train the RENET2 model and conduct their training data expansion using the other 500 abstracts. They further used the model trained on 1,000 abstracts to construct 500 full-text articles. |
| **Drug-gene-mutation** |  |  |  |  |  |
| N-ary [24] | - | - | 3,462 triples; 137,469 drug-gene relations; 3,192 drug-mutation relations; | Cross-sentence | Authors use distant supervision to construct a cross-sentence drug-gene-mutation RE dataset. They use 59 distinct drug-gene-mutation triples from the knowledge bases to extract 3,462 ternary positive relation triples. The negative instances are generated by randomly sampling the entity pairs/triples without interaction. |
| **Event extraction** |  |  |  |  |  |
| BioNLP ST 2009 GE [25] | 1,200 abstracts | - | 13,623 events | Sentence | As the first BioNLP shared task, it aimed to define a bounded, well-defined bio event extraction task, considering both the actual needs and the state of the art in bio-TM technology and to pursue it as a community-wide effort. |
| BioNLP ST 2011 ID [26] | 30 full-texts | 12,740 entities | 4,150 events | Sentence | The ID task focuses on the functions of a class of ubiquitous signaling systems in bacteria, and includes the molecular mechanisms of infection, virulence, and resistance. They extend the BioNLP'09 Shared Task (ST'09) event representation for the ID dataset, which consists of30 full-text publications on infectious diseases. |
| BioNLP ST 2011 EPI [26] | 1,200 abstracts | 15,190 proteins | 3,714 events | Sentence | The EPI task aims to extract the events regarding chemical modifications of DNA and proteins related to the epigenetic control of gene expression. |
| BioNLP ST 2011 REL [26] | 1,210 abstracts | 14,966 proteins | 2,834 relations | Sentence | In contrast to these two application-oriented main tasks, the REL task generally seeks to support extraction by separating challenges relating to part-of relations into a subproblem that independent systems can address. Data for the supporting task REL was created by extending previously introduced GENIA corpus annotations. |
| BioNLP ST 2011 GE [27] | 1,210 abstracts; 14 full-text | 21,616 proteins | 18.047 events | Sentence | The GENIA event (GE) task follows the task definition of BioNLP shared task (ST) 2009, which is briefly described in this section. BioNLP ST 2011 took the role of measuring the progress of the community and generalization IE technology to the full papers. |
| BioNLP ST 2013 CG [28] | 600 abstracts | 21,683 entities | 17,248 events; 917 relations | Sentence | The Cancer Genetics (CG) corpus contains annotations of over 17,000 events in 600 documents. The task addresses entities and events at all levels of biological organization, from the molecular to the whole organism, and involves pathological and physiological processes. |
| BioNLP ST 2013 PC [28] | 525 abstracts | 15,901 entities | 12,125 events; 913 relations | Sentence | The pathway curation (PC) task aims to develop, evaluate and maintain molecular pathway models using representations such as SBML and BioPAX. The PC task stands out in particular in defining the structure of its extraction targets explicitly regarding major pathway model representations and their types based on the Systems Biology Ontology, thus aligning the extraction task closely with the needs of pathway curation efforts. The PC corpus over 12,000 events in 525 documents. |
| BioNLP ST 2013 BB [29] | 131 abstracts | 5183 entities | 2312 events | Sentence | The Bacteria Track tasks aim to demonstrate that the BioNLP community is well-grounded to accompany the progress of Microbiology research. BB targets ecological information for a large spectrum of bacteria species. |
| BioNLP ST 2013 GRN [29] | 201 sentences | 917 entities | 819 events | Sentence | The GRN task targets biological processes and whole cell models. The GRN task's goal is to extract a regulation network from the text. They defined six interaction types for the GRN regulation network representing the whole range of effect and mechanism regulation types |

| BioNLP ST 2013 GRO [29] | 300 abstracts | 11,819 entities | 5,241 events | Sentence | The Gene Regulation Ontology (GRO) task aims to evaluate systems for extracting complex semantic representation in gene regulation domain. |
|---|---|---|---|---|---|

## Acknowledgements

This research is supported by the Intramural Research Programs of the National Institutes of Health, National Library of Medicine and partially supported by the NIH grant 2U24HG007822-08 to CNA.

Conflict of Interest: none declared.